\documentclass[10pt,twocolumn,letterpaper]{article}

\usepackage{iccv}
\usepackage{times}
\usepackage{epsfig}
\usepackage{graphicx}
\usepackage{amsmath}
\usepackage{amssymb}

\usepackage[pagebackref=true,breaklinks=true,letterpaper=true,colorlinks,bookmarks=false]{hyperref}

\iccvfinalcopy 

\def\iccvPaperID{33} 

\ificcvfinal\fi
\begin{document}

\title{Vision-Based Fallen Person Detection for the Elderly}

\author{Markus D. Solbach and John K. Tsotsos \\
Department of Electrical Engineering and Computer Science\\
York University, Canada\\
{\tt\small \{solbach, tsotsos\}@eecs.yorku.ca}
}

\maketitle
\thispagestyle{empty}

\begin{abstract}
Falls are serious and costly for elderly people. The Centers for Disease Control and Prevention of the US reports that millions of older people, 65 and older, fall each year at least once. Serious injuries such as; hip fractures, broken bones or head injury, are caused by $20\%$ of the falls. The time it takes to respond and treat a fallen person is crucial. With this paper we present a new , non-invasive system for fallen people detection. Our approach uses only stereo camera data for passively sensing the environment. The key novelty is a human fall detector which uses a CNN based human pose estimator in combination with stereo data to reconstruct the human pose in 3D and estimate the ground plane in 3D. Furthermore, our system consists of a reasoning module which formulates a number of measures to reason whether a person is fallen. We have tested our approach in different scenarios covering most activities elderly people might encounter living at home. Based on our extensive evaluations, our systems shows high accuracy and almost no miss-classification. To reproduce our results, the implementation is publicly available to the scientific community.
\end{abstract}
\section{Introduction}
Falls are one of the most dangerous situations for elderly people at home \cite{Rougier, Wild1981, Scott2010} and leading reason for injury-related hospitalization. \cite{Stevens2006} reports that annually \$31 billion dollar are spent on direct medical costs for fall injuries in the US. Besides preventing falls in the first place, it is crucial to detect falls as fast as possible. \cite{Wild1981, Marquis-Faulkes2005} point out that $50\%$ of those who lay on the floor for more than an hour, died within six months after the fall, even though the person did not suffer any direct phyical injury. In such cases, death was usually from bronchopneumonia, dehydration or hypothermia. The observation of \cite{Doughty2000} claims that lying down for a long time is as relevant to decreasing the possibility of survival as a broken bone. In other words, it is of great interest to detect a fall immediately. Furthermore, the probabilty of the recurrence of a fall increases dramatically after a person falls for the first time. Unfortunately, a fallen person is likely to fall again with double the chances as \cite{OLoughlin1993} shows. The health industry has a big demand for assistive technology for elderly in fall detection. \cite{Yu2008} explains that
\begin{quote}``with the rapid growth of the population of the elderly in the world, the demand for the healthcare system is increased accordingly. At the same time, advances of sensor, camera, and computer technologies make such development feasible."
\end{quote}
The Kellogg international working group on the prevention of falls in the elderly, defines a fall as ``unintentionally coming to ground, or some lower level not as a consequence of sustaining a violent blow, loss of consciousness, sudden onset of paralysis as in stroke or an epileptic seizure"\cite{GibsonM.J.S.AndresR.O.KennedyT.E.Coppard1987}. Therefore, different types of falls can be distinguished. \cite{Yu2008} describes falls either as \textit{falls from sleeping}, for instance falling of a bed, \textit{falls from sitting}, for instance falling from a chair, sofa or similar and lastly, \textit{falls from walking or standing}, for instance tripping over the edge of a carpet and other floor unevenness. One of the most descriptive characteristics of a fall is the sudden reduction in the height of the head \cite{Rougier,Yu2008,Yu2012} or a person with their head close to the ground plane longer for a certain time can be considered as fallen \cite{Yu2012}. In recent years, modern technology has advanced and several systems were developed to provide ageing in place for the elderly. In this work, we will focus on fall detection systems for the elderly. In general, fall detection systems can roughly be categorized into two groups.
\\\textit{Group 1: Wearable Sensor --} Wearable sensor based devices such as the Zenio system of Vitaltronics, the Fall Detection System of Galaxy Medical Alert Systems and AutoAlert from Philips. Most of such systems make use of accelerometers to detect abnormal changes in height orientation to a horizontal position and velocity to trigger an alarm \cite{LindemannUlrich2005,Zhang2006}. As reported, these systems are generally accurate in detecting falls, but they all have one drawback: The person needs to wear them, which raises two problems. Firstly, the person might forget to wear the sensor. Secondly, the person is reluctant to wear it, due to wounded vanity or discomfort. The systems presented so far are able to generate an alarm automatically based on a fall event. A subcategory of wearable sensors exist which are purely manual. The main difference is that such systems do not detect a fall, they are operated entirely by the elderly and once someone is in need, a button can be pushed to ask for help. Besides the already mentioned drawbacks of wearable sensors, these devices do not cope with situations where the person is not able to operate the device, for instance due to unconsciousness, bone fractures or other injuries.
\\\textit{Group 2: Video-Based --} Video-Based systems offer many advantages over systems based on wearable sensors. The elderly will not worry about using the device. Once installed and set up, the video-based solution becomes part of their home. No physical contact is needed with the elderly. Cameras monitor the home passively and without being invasive, in preparation to detect a fall event. In case of a fall, a video snipped or live stream can be send to a family member or service provider for elderly care to reassure the fall event before requesting medical assistance. However, video-based systems come with their own drawbacks which are important to mention. \cite{Williams2007} points out the privacy concerns that comes along with using video to detect a fall event. The bathroom, especially is a crucial location because it is a high fall-risk location and a place for private situations, which might make it uncomfortable for the elderly being filmed. The elderly needs to be fully aware of the situation and it is important to take privacy very seriously. The video must not be available to any form of a third party, except for the known parties. The system needs to have the option to make the video live stream anonymous, for instance, blur out the fallen person and exposed private parts. Another important limitation of video-based systems is the coverage. A fall can be detected only where the camera is looking. This can be overcome by either installing cameras in each room or having a camera mounted on an autonomous robot. For the sake of completeness, a video-based system needs to be robust to occlusion and lighting changes. Video-Based systems are presented in detail in Section \ref{sect:related}.
\\With this paper we present a novel video-based fall \mbox{detection} system. The key contributions are 3D human key point estimation based on a state-of-the-art 2D human Pose estimation \cite{Cao2016} and stereo depth data, the introduction of multiple measures on which to reason whether a person is fallen, a versatile system that can be either installed in a stationary, multi-camera setup or on an active observer like an autonomous robot, wide range of experiments in home-like scenarios and lastly the release of our source code for reproducibility.
\\The remainder of this paper is organized as follows. Section \ref{sect:related} gives an overview of existing video-based fall \mbox{detection} systems to provide a perspective on the state-of-the-art. Section \ref{sect:approach} presents our approach in detail. Section \ref{sect:experiments} shows results from our mainly home-like experiments including different scenarios such as a fallen person covered in bedsheets, fallen off a couch, tripped over the edge of carpet and many more. Section \ref{sect:conclusion} concludes our work and gives an outlook on future work.

\section{Related Work}\label{sect:related}
In this Section we present related work in fallen person detection using vision. Based on the corpus of the work, we decided to split this section into two Subsections. Section \ref{subsect:systems} presents complete systems which are designed specifically to detect fallen persons. Section \ref{subsect:pose} gives an overview of human pose estimators, which is important since our \mbox{approach} uses the human pose to reason whether a person is fallen. For completeness, the reader is pointed to the \mbox{survey} papers \cite{Noury2007,Willems2009,Williams2007,Zhang2015} which provide a comprehensive overview of fall detection systems.

\subsection{Vision-Based Fall Detection System}\label{subsect:systems}
Most of the vision-based fall detection system uses some form of background subtraction to firstly distinguish between the environment and person and secondly reason whether a person is fallen \cite{Lee2005,Toreyin2006,Rougier2011,Cucchiara,Hazelhoff,Belshaw2011,Vaidehi2011}.
\\One example of such a system is an automated fall detection system proposed by \cite{Lee2005} using a ceiling mounted camera. With background subtraction full body images were detected and characterized with the vectorized silhouette of their shape. It is reported that ``tucked" falls could not be detected and occlusion was not handled either. Other shortcomings were that this system was only able to monitor one person and the performance suffered in the presence of other objects, such as walking sticks. Another system based on background subtraction is presented in \cite{Toreyin2006}. The moving human is detected based on motion. The size and geometric shape of the bounding box is used to determine whether a person is fallen or not. The evaluation is performed on rather simplistic scenarios with minimal clutter and no occlusions. The problem of occlusion was addressed in \cite{Cucchiara} using a multi-camera setup and assuming that the fallen person is unoccluded in at least one camera frame. This approach uses an appearance-based object tracker \mbox{using} color and geometry cues to identify people. Different human postures, such as standing, crawling, sitting and lying were classified with a Hidden Markov Model. Drawbacks are that the system assumes that the person is unoccluded in the camera frame and not many results were presented by the authors. A similar multi-camera setup is proposed by \cite{Williams2007}. The detection of a fall is rather simple by extracting the moving person from the background and calculating the aspect ratio of height-width to determine if a person is fallen. \cite{Vaidehi2011,Charfi2012} are other examples using the aspect ratio of height-width of the detected bounding box. A system using two fixed cameras and a Gaussian Mixture Model (GMM) is presented by \cite{Hazelhoff}. Using GMM background subtraction, foreground objects are detected and features, such as direction of the main axis of the body and the ratio of the $x$ and $y$ direction variances, are calculated. The authors use these features to differentiate between walking and falling. A head tracker using skin colour cues is incorporated to further improve the results of the classification. This work claims a fall detection rate of $85\%$ on over $4,000$ images in scenarios containing furniture, shadows and difficult lighting conditions. Using a single wide angle overhead camera, \cite{Belshaw2011} proposes another vision-based fall detection system based on foreground detection using background subtraction. Falls are detected using silhouette features in combination with three classifiers. The authors use logistic regression, neural network and support vector machines to classify fall and no-fall events. Out of the three classifiers, the neural network performed the best with a True Positive Rate of $92\%$ and a False Positive Rate of $5\%$. Other approaches using head tracking are presented in \cite{Rougier,Rougier2011}. The problem of occlusion of the body of a person through self-occlusion or occlusion through other objects like a blanket, is addressed by tracking the head. An important assumption is that the head of the subject will undergo minimal occlusions. A monocular camera is used to track the ellipse of the subject with a particle filter tracker. The tracking output was a 3D trajectory of the head and was classified with a simple binary classifier to indicate a fall. Unfortunately, this system is different in two ways from the other presented systems. Firstly, it is not automatic. The ellipse of the test subject were initialized manually. Secondly, this system detects the action of a fall and not a fallen person. However, this paper is worth mentioning since our approach uses cues of the head position as well.

\subsection{Human Pose Estimation}\label{subsect:pose}
We believe that identifying the human pose to determine whether a person is fallen is crucial. The corpus of work is significant in estimating the human pose and recent advances in deep learning enable new possibilities in estimating the human pose. This Section will first present vision-based fall detection systems that use the human pose as a cue and the remainder of this Section presents current state-of-the-art human pose estimators based on deep learning.
\\A healthcare robot is developed in \cite{Wang2012} that detects lying poses. Each image was segmented to reduce the image search space. As most systems in Section \ref{subsect:systems}, the authors use foreground detection to reduce the image search space. Each foreground mask is used to perform pose estimation using the method of articulated bodies of \cite{Ramanan2007}. To train the system the authors created the \textit{FT Lying Person Dataset} which consists of indoor images obtained from the internet, and the PASCAL 09 \cite{Everingham2010a} dataset for negative examples. Unfortunately, the \textit{FT Lying Person Dataset} is not publicly available. Based on the given example images, it is not clear how this system performs with partial occlusion. Furthermore, the authors mention that their system does not use any spatial information and detecting the ground plane could help reject false alarms. \cite{Yu2012} detects the human pose based on an ellipse fitted to the detected human and the posture is estimated by calculating ellipse axes. A support vector machine is used to classify different human poses. In comparison to \cite{Wang2012} the pose and its relation to the ground is used to detect a fallen person. Work based on 3D point clouds using the Microsoft Kinect sensor is addressed in \cite{Volkhardt2013}. The approach uses point cloud clustering to detect different limbs and for spatial relations, the ground plane is segmented. A feature vector using geometrical features, local surface normals, fast point features histograms and surface-entropy is used. As classification methods the authors use different methods such as support vector machines, random forests and AdaBoost. Even though this work addresses the problem of partial occlusion, the test scenarios are fairly simple like white walls with minimal occlusion and the fallen person is to be assumed to be always on the floor surface. Another approach using the Microsoft Kinect sensor is presented by \cite{Mastorakis2014}. The 3D bounding box, of a human is calculated using background subtraction and the depth data. Based on velocity features of the 3D bounding box falls and non-falls were detected. Also, this work deals with the event of a person falling and not a fallen person. The work by \cite{Mundher2014} is the closest to our proposed approach. The human skeleton is used to detect a fallen person. Joint information of the skeleton is achieved from the Microsoft Kinect sensor. To reason whether a person is fallen the authors use the height of the joints with respect to the detected ground. The authors present samples with a rather dominant presence of a human, with no clutter or occlusion.
\begin{figure}[t]
\begin{center}
\includegraphics[width=0.8\linewidth]{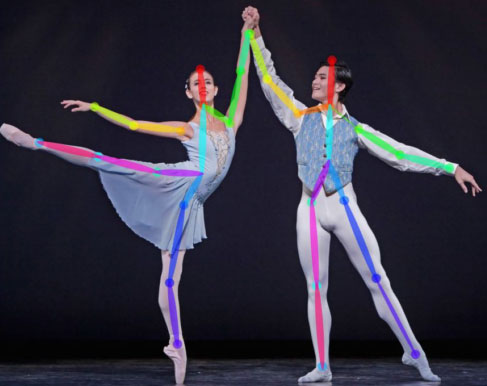}
\end{center}
   \caption{Example of a 2D Human pose estimation using \cite{Cao2016}.}
\label{fig:openpose}
\end{figure}
\\In recent years, deep learning became more and more popular. Most of modern human pose estimators rely on deep learning. Rich training datasets are publicly available \cite{JiaDeng2009,Lin2014a,Andriluka2014} to develop and train new deep learning approaches, such as human pose estimation. \textit{MPII Human Pose} \cite{Andriluka2014} for instance, is a dataset specifically designed for 2D human pose estimation containing $25,000$ images with over $40,000$ people with annotated body joints. Many human pose estimators make use of this rich dataset and achieve impressive results with average accuracies around $90\%$ \cite{Ning2017,Bulat,Newell2016,Cao2016}. All of the mentioned deep learning approaches work on single color images and estimate 2D human poses. An example illustration can be seen in Figure \ref{fig:openpose}. \cite{Cao2016} performs with $91.5\%$ accuracy on the  \textit{MPII Human Pose} Dataset best as of the creation of this paper. In comparison to other approaches, besides the incredible performance, this approach is able to correctly detect up to $19$ people with $8.8$ FPS. The source code is publicly available with well written documentation that makes it easy to use. A common drawback of deep learning approaches is the costly demand on hardware the accelerate these algorithms. Using NVIDIA's CUDNN, the implementation based on Caffe \cite{Jia} needs only a low to mid-range GPU with $1.5$ GB memory. All these benefits make \cite{Cao2016} very attractive as a 2D human pose estimator for our work.
\\However, none of the presented deep learning approaches detect fallen person, they only estimate the human pose. It is the task of this paper to introduce a system extending generic 2D human pose estimators to reason whether a person is fallen.

\section{Approach}\label{sect:approach}
In this Section we present our work. As shown in Section \ref{subsect:pose}, deep learning approaches achieve good results in detecting the human pose in 2D on a single color image. The novelty of this paper is to use a generic 2D human pose estimator in combination with depth information to estimate 3D human key points, calculate the ground plane in 3D and reason by using multiple measures as to whether a person has fallen. Furthermore, we present a versatile system that can be either installed in a stationary, multi-camera setup or on an active observer like an autonomous robot.
\begin{figure}[t]
\begin{center}
\includegraphics[width=0.8\linewidth]{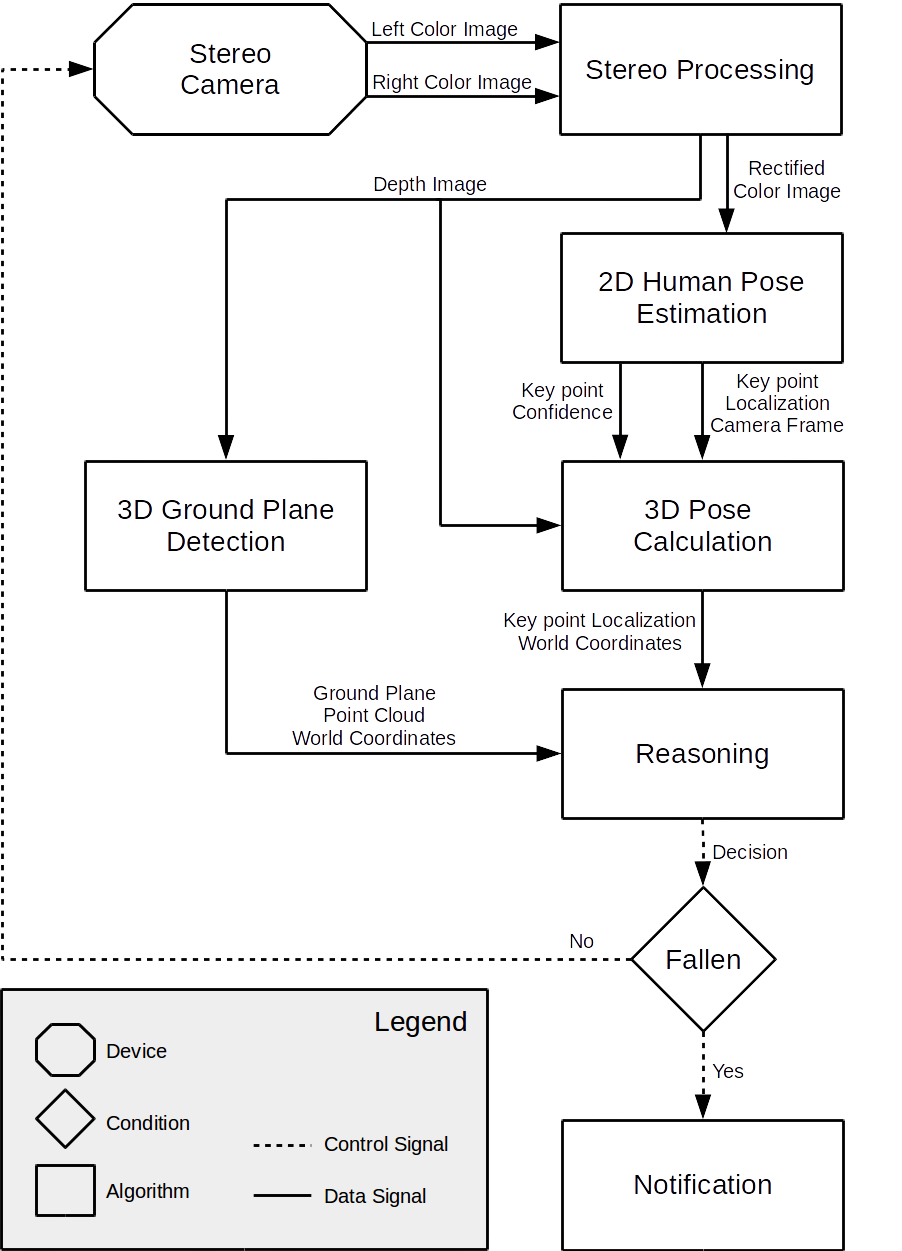}
\end{center}
   \caption{Overview of our proposed system.}
\label{fig:overview}
\end{figure}
\\Figure \ref{fig:overview} illustrates the overview of our system. We use a \textit{Stereo Camera} to sense the environment and calculate the depth image. In comparison to other depth sensing approaches like \cite{Mastorakis2014,Mundher2014,Volkhardt2013} which use a Microsoft Kinect, we use a purely non-invasive and passive sensing approach. Step \textit{Stereo Processing} rectifies left and right color image and calculates the depth image. \textit{2D Human Pose Estimation} is using the deep learning approach of \cite{Cao2016} to estimate the human pose in 2D. This step provides us with pixel coordinates for each detected key point and a confidence level. For this step the left color image is used, since we perform any calculation with respect to the left camera later. \textit{3D Pose Calculation} uses the camera matrix, depth image and key point information of the previous step to project the key points into 3D World-Coordinates. \text{3D Ground Plane Detection} uses the depth image as well to calculate the ground plane in 3D which is important for the \text{Reasoning} step. In the \textit{Reasoning} step our system applies a number of measures to determine if a person is fallen. This decision is contingent upon the ground plane, 3D pose and confidence levels of each detected key point. Dependent on the result of the \textit{Reasoning} step, the system will either perform a \textit{Notification} if a fall was detected or start over with \textit{Stereo Camera}. The whole system is implemented in Robot Operating System (ROS) and is publicly available\footnote{\url{https://github.com/TsotsosLab/fallen-person-detector}}.

\subsection{3D Human Pose Calculation}\label{subsect:3Dpose}
For the 2D human pose estimation we use the approach of \cite{Cao2016}. Based on our empirical evaluations, \cite{Cao2016} performs not only best on the \textit{MPII Human Pose} Dataset, it also performed best on a small dataset containing images from the internet showing only fallen people. \cite{Cao2016} derives its convolutional neural network partially from VGG-19 \cite{Simonyan2015a} and it is fine-tuned on \textit{Microsoft's COCO} dataset \cite{Lin2014a} and \textit{MPII Human Pose} dataset \cite{Andriluka2014}. Furthermore, this approach is unique in a way that it presents an explicit nonparametric representation of key point association that encodes position and orientation of a human limb. The latter, called part affinity fields for parts association, is important to differentiate detections between multiple persons.
\\We use the Microsoft COCO human description for our work which describes the human body pose with 19 key points in 2D with confidence levels for each key point. Key points are; right and left ears, eyes, shoulders, elbows, wrists, hips, knees, ankles and nose, neck and chest. To our knowledge, there was no ROS enabled implementation of \cite{Cao2016}. Our ROS-Wrapper for \cite{Cao2016} is published for the scientific community with this paper\footnote{\url{https://github.com/TsotsosLab/openpose-ros}}. If step \textit{2D Human Pose Estimation} detects at least one person with a normalized average confidence over all key points above $60\%$ the system continues with \textit{3D Pose Calculation}. Equation \ref{equ:normalized}) illustrates the calculation of the normalized average confidence
\begin{equation}\label{equ:normalized}
\begin{split}
\Lambda = \frac{1}{K} \cdot \sum_{k=1}^{K} \lambda_k,
\end{split}
\end{equation}
\\where $K$ is the number of detected key points and $\lambda_k$ denotes the confidence of the $k-th$ key point.
\\To obtain the 3D pose of 2D key points detection we incorporate the depth image and camera matrix. Equation \ref{equ:3DCalcu} shows the calculation to achieve the 3D world coordinates $(X^{i,j}_w, Y^{i,j}_w, Z^{i,j}_w)$ for a given key point $k$ at pixel coordinate $(x^{i,j}_c, y^{i,j}_c)$

\begin{equation}\label{equ:3DCalcu}
\begin{split}
\Phi_{k} = \left[ \begin{array}{ll}
         Z^{i,j}_w = d^{i,j}\\
         X^{i,j}_w = (x^{i,j}_c - c_x) \cdot \frac{d^{i,j}}{f_x}\\
         Y^{i,j}_w = (y^{i,j}_c - c_y) \cdot \frac{d^{i,j}}{f_y}
        \end{array} \right],
\end{split}
\end{equation}
\\where $c_x$ is the principal point in $x$ direction and $c_y$ the principal point in $y$ direction, $f_x, f_y$ denotes the focal length and $d^{i,j}$ is the depth at pixel position $(i,j)$. At this point we assume that depth image and rectified color image are aligned to each other, such that we can simply read the depth of key points given the $(x,y)$ pixel coordinates of this key point. This equation is executed for each key point. After this step we have all key points in 3D world coordinates stored in $\Phi$ which is important for further steps.

\subsection{Ground Plane Detection}\label{subsect:ground}
For our work, the ground plane detection is crucial. Other authors like \cite{Wang2012,Volkhardt2013} incorporated ground plane information and showed good results results. \cite{Everingham2010a} specifically points out in their conclusion that detecting the ground plane would result in less false positive detections. We formulate the following requirements for a ground plane detection algorithm in the scenario of a fallen person detection system:
\begin{itemize}
    \item \textit{fast --} Processing time is important for our overall system. Our goal is to have a system that can also be deployed on an autonomous robot, which reacts to its environment without significant computational effort.
    \item \textit{reliable --} The ground plane detection is the backbone of our fallen person detection system, hence the reliability of our system depends on the ground plane detection
    \item \textit{slope tolerant --} A person might fall in different scenarios, they might end up collapse next to a bed and end up lying half on the bed and half on the ground covered in a blanket or they might collapse on a ramp like structure. Therefore, the ground plane detection needs to be tolerant of the slope to accurately detect such cases as well as cases in which the person lies on the flat ground.
\end{itemize}  
Based on our research, a floor detection algorithm that satisfies all three requirements is presented by \cite{Zhang2003}. Initially implemented for Airborne LIDAR (LIght Detection And Ranging) data, this algorithm also works well with other point cloud data. The algorithm is based on the commonly used idea of the mathematical morphology filter applied to grayscale images to remove non ground objects.
\begin{figure}[t]
\begin{center}
\includegraphics[width=0.8\linewidth]{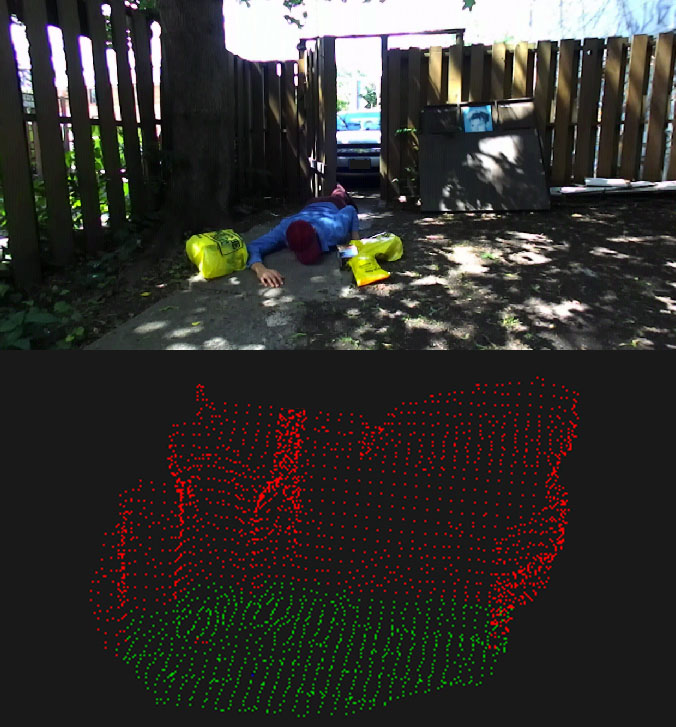}
\end{center}
   \caption{Example 3D Ground Plane Detection. Green is classified as ground objects, red is classified as non ground objects.}
\label{fig:groundfloor}
\end{figure}
It is based on the idea that the elevation of objects, like furniture, is usually higher than those of surrounding ground points. LIDAR or point cloud data can be translated into a grayscale image, where intensity is reflects the elevation, such that objects with different elevation can be identified by the change of gray tone. As described by the authors of \cite{Zhang2003} mathematical morphology can be used to filter LIDAR data. One difficulty is to detect non ground objects of various sizes using a fixed filtering window size. To overcome this problem \cite{Zhang2003} introduces a progressive morphological filter which increases the window size of the opening operation. Furthermore, an elevation difference threshold is introduced which gives us control over the slope tolerance.
\\Our test system uses the Stereolab's ZED stereo camera which produces fairly dense point clouds of the environment. We need to introduce a filtering step to minimize the data we use to detect the ground plane. To fulfill the requirement of being \textit{fast}, we subsample the point cloud to a voxelgrid with leaf sizes of $15cm$. This minimizes, for instance, a point cloud with around $100,000$ points to one with $2,000$ points. Even though the resulting cloud is $98\%$ smaller, characteristic features are preserved since we subsample the cloud and do not blindly discard data. An example is given in Figure \ref{fig:groundfloor}. On the top the input color image is shown and on the bottom the result of the ground plane detection is illustrated. The ground plane detection uses green for ground objects and red for non ground objects. The algorithm detects correctly ground objects. With a closer look, it can be seen that the left shopping bag is partially detected as a non ground object due to its steep slope compared to its neighbors. Due to its flat appearance and little slope, the fallen person is also detected as a ground object.

\subsection{Reasoning}\label{subsect:reason}
\begin{figure}[t]
\begin{center}
\includegraphics[width=0.8\linewidth]{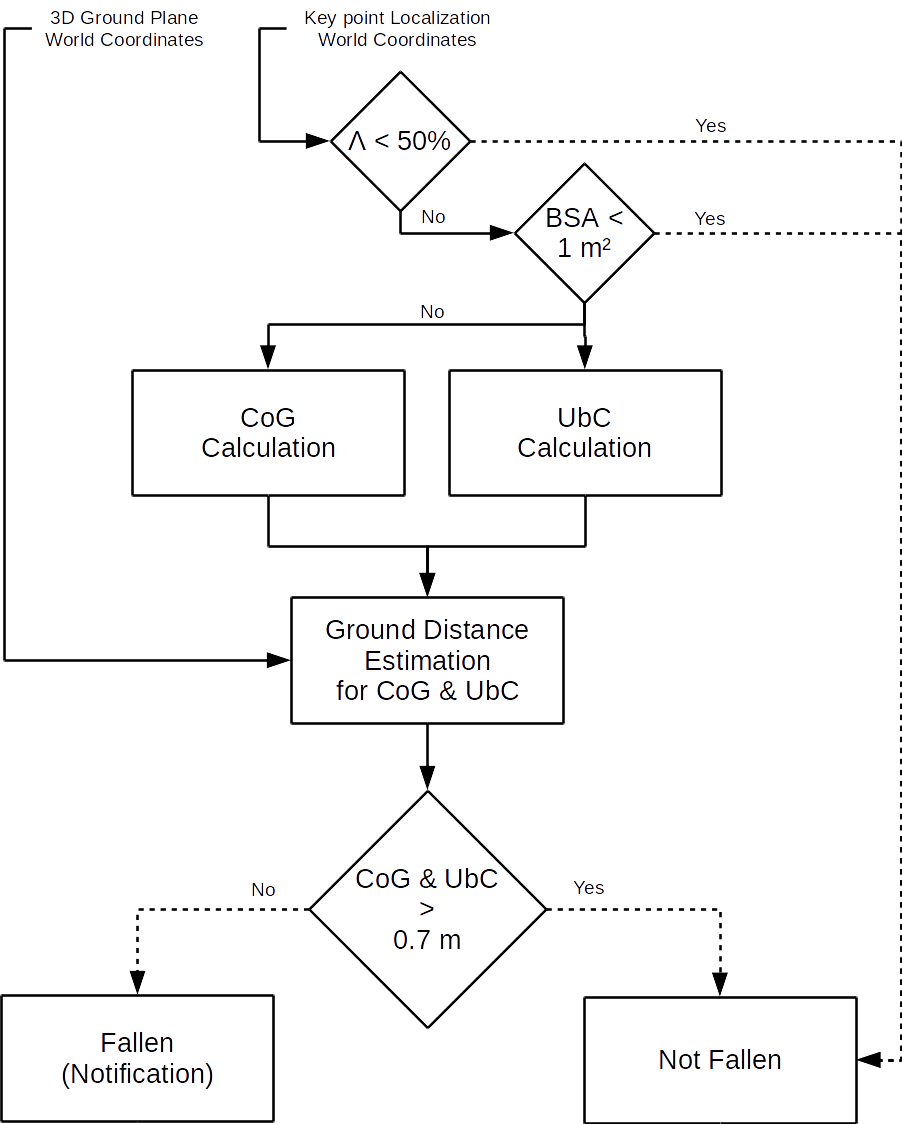}
\end{center}
   \caption{Overview of the reasoning process of our proposed algorithm. See Figure \ref{fig:overview} where \textit{3D Ground Plane} and \textit{Key point Localization} comes from and where \textit{Fallen} and \textit{Not Fallen} goes to.}
\label{fig:reasoning}
\end{figure}

At this phase of the system we have gathered information to reason if a person is fallen. The \textit{Reasoning} component needs as input the 3D ground plane, 3D location of each detected key point and the confidence of each key point. The output is a decision that either a person is \textit{Fallen} or \textit{Not Fallen}. In the first case an alarm can be triggered, a message can be sent to family members or a short video snippet can be transmitted to an elderly care service provider to further judge the situation. In the other case if no fall is detected, the system obtains the next image and starts the procedure again, as illustrated in Figure \ref{fig:overview}. Figure \ref{fig:reasoning} shows the overview of the reasoning procedure of our algorithm whose components are now described.
\\\textit{$\Lambda$ --} The 2D human pose estimation by \cite{Cao2016} produces estimates with confidence levels as low as $0.0\%$. It is our task to reason about the confidence level in a way to discard false positive detections. Based on thorough empirical studies we found that the average confidence over all detected key points $\Lambda$ needs to be at least $50\%$ to be a valid human pose estimation. A key point counts as detected if its confidence is $>1.0\%$. If $\Lambda < 50\%$ we can consider the detection as a false detection, hence no person is fallen. Another observation of the 2D human pose estimation was that due to the lack of depth information, small, richly detailed textures in close objects were detected as humans. A human observer can identify such false detections easily, since the detected pose looks very condensed and also the positions of the limbs look unusually folded together. However, it is not simple for a machine to identify such false detections.
\\\textit{$BSA$ --} Besides $\Lambda$, we introduce another measure to decide if we have a false detection. Using 3D key point information we are able to estimate the bounding box of the human in 3D. Based on the bounding box we can estimate the width and height of the detected human, which gives us an estimated surface area for this bounding box. To reason whether an estimated surface area is reasonable for a human or not, we looked into the work of \cite{DuBois1989} which introduces an calculation for the body surface area ($BSA$) based on height and weight. Equation \ref{equ:bsa} shows the calculation of the $BSA$ as proposed by \cite{DuBois1989}
\begin{equation}\label{equ:bsa}
\begin{split}
BSA = 0.007184 \cdot W^{0.425} \cdot H^{0.725},
\end{split}
\end{equation}
where $W$ denotes the mass in kg, $H$ height in cm and $BSA$ the surface area in $m^2$.  
Dependent on the age and gender, the average BSA value can differ from $1.4m^2$ to $1.9m^2$ which can be seen by entering appropriate values into Equation \ref{equ:bsa}. For our system, we decided if the detection has an area below $1m^2$ we have likely encountered a false detection, hence no person is fallen. 
\\\textit{$CoG$ --} $CoG$ stands for center of gravity. With this measure we take into account all detected key points and calculate the center of gravity of the detection.
\begin{equation}\label{equ:cog}
\begin{split}
CoG = \frac{1}{K} \cdot \sum_{k=1}^{K} \left( \begin{array}{ll}             
        \lambda^X_k\\
        \lambda^Y_k\\
        \lambda^Z_k
        \end{array} \right)
\end{split}
\end{equation}
Equation \ref{equ:cog} shows the calculation of $CoG$ with $\lambda^X_k$, $\lambda^Y_k$and $\lambda^Z_k$ denoting the $X$-, $Y$- and $Z$-Coordinate of the $k-th$ keypoint detection. Basically, we take the sum of each coordinate from each key point and divide by the number of detected key points.
\\\textit{$UbC$ --} $UbC$ stands for upper body critical. Similar to $CoG$ we calculate the center of gravity but in this case for a subset of the key points. The subset consists of key points related to nose, eyes, ears, neck and shoulders. We chose this subset to emphasize that a person with those body parts close to the ground is in a critical condition and is most likely fallen, especially if the person remains in this posture longer than a certain time, a fall is encountered.   
\\\textit{$Ground~Distance$ --} The remaining part of the reasoning module merges the 3D ground plane with $CoG$ and $UbC$. Using Euclidean distance we compare each 3D point of the ground plane with $CoG$ and $UbC$. If one of them is lower to the ground than $0.7m$ we classify the human as fallen. We chose $0.7m$, since most lying- and seating-accommodation for the elderly are higher than $0.7m$, therefore, if we detect a human below this threshold we most likely detected a fall. We differentiate between $CoG$ and $UbC$ in the case of someone falling down on a bed or couch but legs remain on the bed or couch. In such a case, $CoG$ will not reach the threshold but $UbC$ will.
\begin{figure}[t]
\begin{center}
\includegraphics[width=1.0\linewidth]{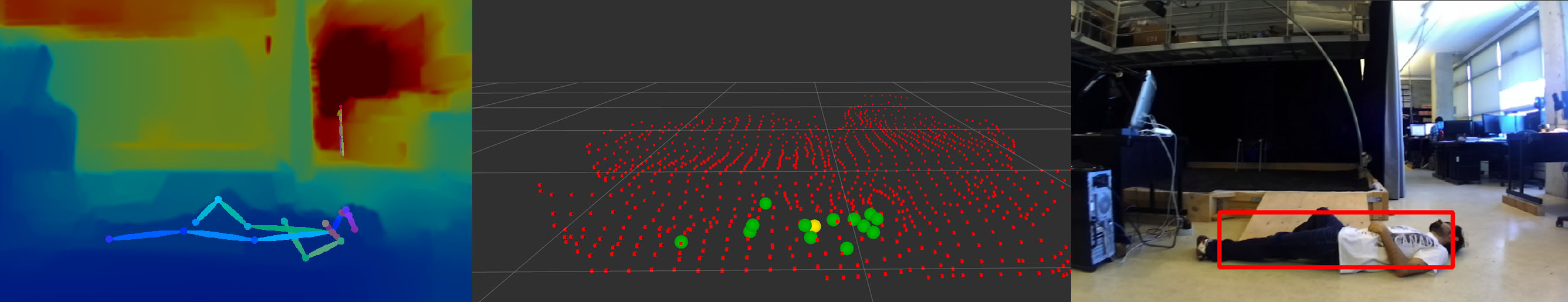}
\end{center}
   \caption{Illustration of our main steps. Left image: depth image visualized with human pose detection. Center image: 3D key points (green), $CoG$ (yellow) and ground plane (red). Right image: Red bounding box to show that a fallen person is detected.}
\label{fig:steps}
\end{figure}
\\Figure \ref{fig:steps} shows an illustration of our main steps. The left image illustrates a color coded depth image (warm colours - far, cold colors - close) with a detected human pose. Circles show key points, while color encoding of limbs and key points are just for visualization. The center image shows the calculated 3D key points in green and $CoG$ in yellow. The ground plane is shown with red cubes. The right image illustrates the final result with a red bounding box indicating that the system detected a fallen person.

\section{Experimental Results}\label{sect:experiments}
In this chapter we present experimental results to highlight strengths and weaknesses in our approach. The system is tested on a Pioneer 3 platform from Mobile Robots with a stereo camera from Stereolabs. The long term goal is an active robot but for now the robot was manually moved. For processing, we used a laptop with NVIDIA GeForce 1070, Intel i7 and 16GB RAM. To our knowledge, no public available data set for fallen person detection exists. We decided, driven from the idea of a fallen person detection for the elderly, to have one test scenario in a home-like environment, covering 6 different rooms and another test scenario in an office-like environment.
\begin{figure}[t]
\begin{center}
\includegraphics[width=1.0\linewidth]{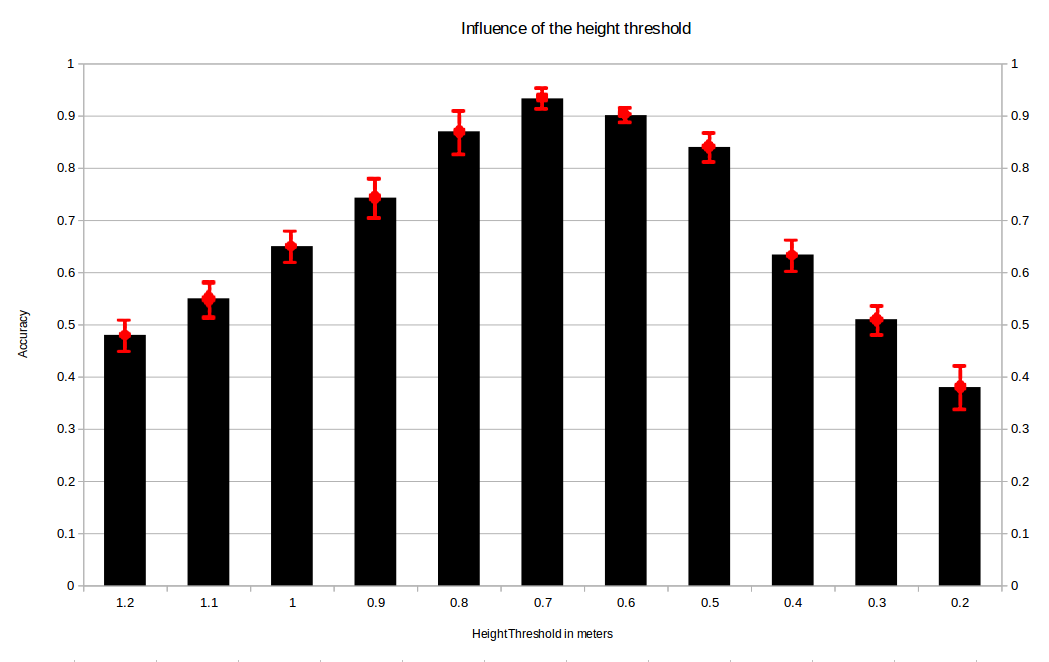}
\end{center}
   \caption{Graphical plot of how the height threshold influences the accuracy of the algorithm.}
\label{fig:height}
\end{figure}
\\Overall, we were able to achieve good results with our novel approach of detecting fallen person. Evaluation on how the height threshold (Figure \ref{fig:reasoning}) in the reasoning module influences the performance, can be seen in Figure \ref{fig:height}. The plot shows accuracy against height threshold in meters. Based on our results we can say that if the threshold is set too high, we detect too many false positives. For example, kneeling persons are not fallen but have easily a $CoG$ below $1.2m$. On the other side, if the threshold is set too low the system is not able to detect fallen person because we have to keep in mind that the ground plane is only detected around the person and not below the person, due to occlusion. For our test data with in total 50 falls and 50 non falls in 7 different scenarios, we found that a threshold around $0.7m$ gives us best results.

\begin{figure}[t]
\begin{center}
\includegraphics[width=0.8\linewidth]{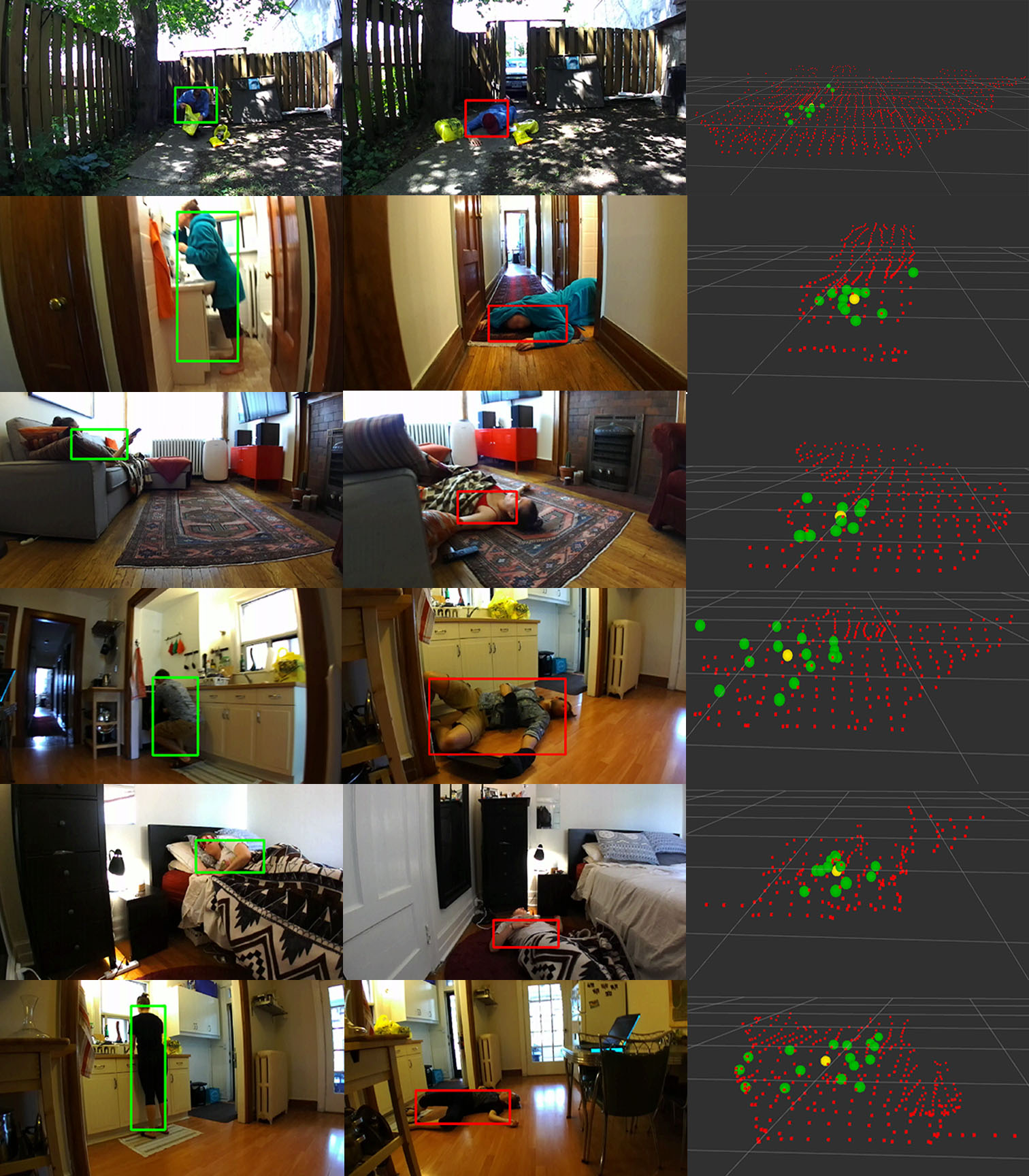}
\end{center}
   \caption{Some images of our experimental results in the home environment. Left shows the scene with no fall detected. The center shows the same scene with a fall detected and the right column shows the corresponding 3D key points with ground plane.}
\label{fig:home}
\end{figure}

\subsection{Home Environment}\label{subsect:home}
One of our two test scenarios is a home environment. We recorded different types of falls in 6 different areas, such as living room, bathroom, hallway, bedroom, kitchen and backyard. Some results are presented in Figure \ref{fig:home}. The left column shows all six areas with a nonfallen person. The middle column shows a fallen person, and the right column visualizes the corresponding ground plane detection (red) and 3D key points (green) with $CoG$ (yellow). The algorithm is reliable even with hard cases, such as partially occluded or a fallen person facing down. In row one we see a scenario where the person is kneeling which can be considered to be very close to fallen but our system reasoned correctly as not fallen. In row two the fall is compared to other fall scenarios reasonably complicated. The person is only half visible and wears the hood of a bathrobe which leaves our system with only a few cues to decide. Another strength of our system can be seen in the third row. Even though the camera only observes very little of the person, our system is able to decide correctly whether the person is fallen. Last but not least we want to point out the fifth row. The system is able to detect a sleeping person wearing its eye covers as not fallen and the same person covered in a blanket on the ground as fallen. In this environment we recorded 30 falls and 30 non falls. Our system achieved a true positive rate of $0.933$.

\subsection{Office Environment}\label{subsect:office}
\begin{figure}[t]
\begin{center}
\includegraphics[width=0.8\linewidth]{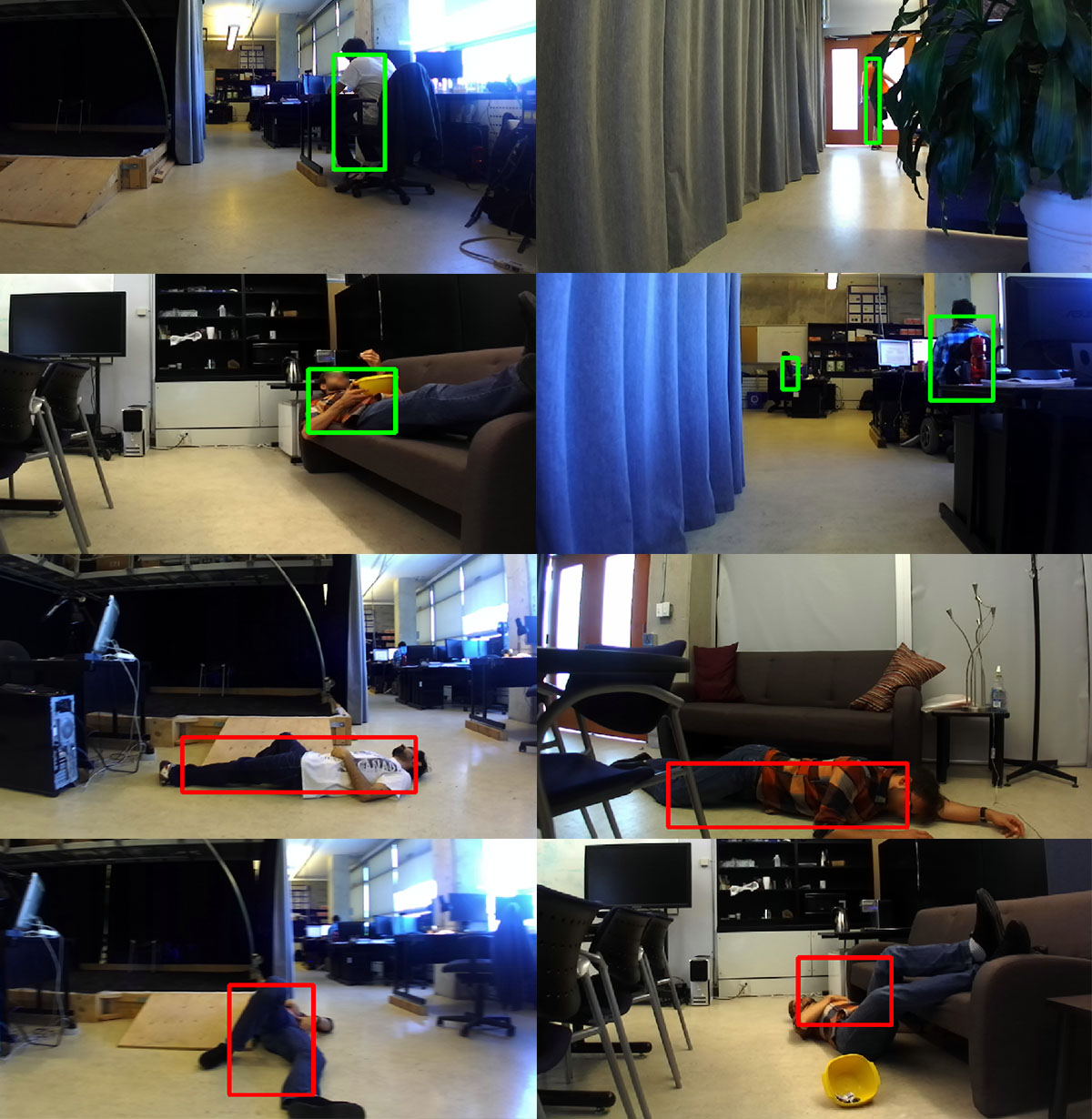}
\end{center}
   \caption{Some images of our experimental results in the office environment.}
\label{fig:office}
\end{figure}
The second test scenario covers 20 falls and 20 non falls in an office environment. In this environment our system achieved a true positive rate of $0.912$. In row two and column two it can be seen that our system is able to process multiple persons at the same time. Row four and column four presents a difficult case in which the person collapsed on a couch and the upper body slid down but the legs remained on the couch. Due to $UbC$ the system was able to detect that this person has fallen. The importance of a ground plane detection with slope can be seen in row four column one. The person is lying half on a ramp. Our system detected the ramp as ground plane, hence the person is fallen.
\begin{figure}[t]
\begin{center}
\includegraphics[width=0.8\linewidth]{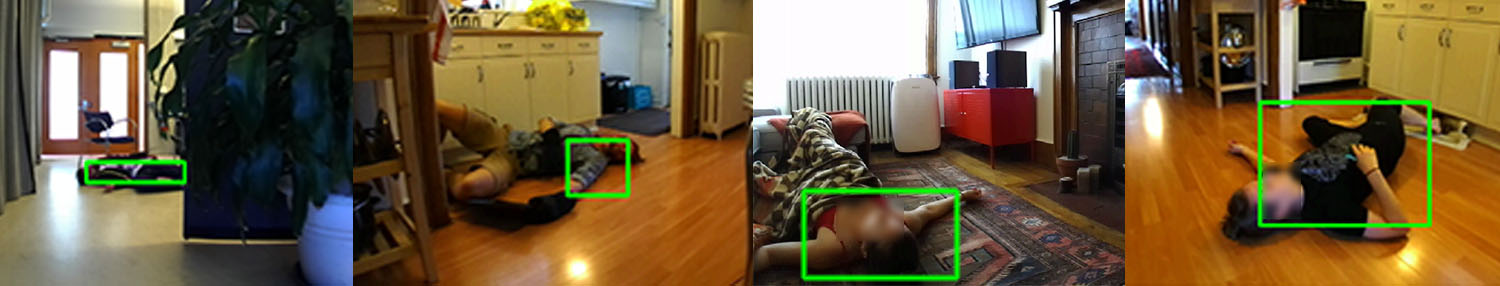}
\end{center}
   \caption{Illustration of some false positive detections. False positives are mainly due to reflections which results in lack of ground data.}
\label{fig:false}
\end{figure}
Lastly, we want to present some false positives in Figure \ref{fig:false}. In all cases the system was not able to directly determine correctly that the person was fallen. Only after the robot was moved to a different position, so that the person was recorded from a different angle, the system was able to detect the person correctly. However, most false positive cases are due to lack of ground plane information which can be explained with reflections on the ground where no depth can be estimated.

\section{Conclusion}\label{sect:conclusion}
In conclusion, we have presented a novel fallen person detector based on stereo vision. We made use of a state-of-the-art 2D human person detection CNN (we published a ROS-Wrapper), extended it to achieve 3D key point information, we made use of ground plane information, formulated a number of measures to reason whether a person is fallen and made our system publicly available for reproducibility. In two large test scenarios, our approach shows outstanding results and was able to achieve accuracies above $91\%$. For future work we want to extend our approach with an autonomous robot that actively observes the scene to decrease false positives and expand and publish our dataset for fallen person detection.

\section{Acknowledgment}\label{sect:acknowledgment}
We would like to thank every person volunteering in the test scenarios as fallen person. This research was supported by several sources for which the authors are grateful: the NSERC Canadian Field Robotics Network (2016-0157), the Canada Research Chairs Program (950-219525), and the Natural Sciences and Engineering Research Council of Canada (RGPIN-2016-05352).

{\small
\bibliographystyle{ieee}
\bibliography{/home/markus/ownCloud/9_literature/mendeley/bibtex/library}
}

\end{document}